# A Systematic Approach to Artificial Agents

Mark Burgin[1] and Gordana Dodig-Crnkovic[2]

[1] Department of Mathematics, University of California, Los Angeles, California, USA,
[2] School of Innovation, Design and Engineering, Mälardalen University, Sweden,

**Abstract.** Agents and agent systems are becoming more and more important in the development of a variety of fields such as ubiquitous computing, ambient intelligence, autonomous computing, intelligent systems and intelligent robotics. The need for improvement of our basic knowledge on agents is very essential. We take a systematic approach and present extended classification of artificial agents which can be useful for understanding of what artificial agents are and what they can be in the future. The aim of this classification is to give us insights in what kind of agents can be created and what type of problems demand a specific kind of agents for their solution.

## 1  Introduction

Agents are advanced tools people use to achieve different goals and to various problems. The main difference between ordinary tools and agents is that agents can function more or less independently from those who delegated agency to the agents. For a long time people used only other people and sometimes animals as their agents. Developments in information processing technology, computers and their networks, have made possible to build and use artificial agents. Now the most popular approach in artificial intelligence is based on agents.

Intelligent agents form a basis for many kinds of advanced software systems that incorporate varying methodologies, diverse sources of domain knowledge, and a variety of data types. The intelligent agent approach has been applied extensively in business applications, and more recently in medical decision support systems [12, 13] and ecology [15]. In the general paradigm, the human decision maker is considered to be an agent and is incorporated into the decision process. The overall decision is facilitated by a task manager that assigns subtasks to the appropriate agent and combines conclusions reached by agents to form the final decision.

## 2  The Concept of an Agent

There are several definitions of intelligent and software agents. However, they describe rather than define agents in terms of their task, autonomy, and communication capabilities. Some of the major definitions and descriptions of agents are given in Jansen [7].

1. Agents are semi-autonomous computer programs that intelligently assist the user with computer applications by employing artificial intelligence techniques to assist users with daily computer tasks, such as reading electronic mail, maintaining a calendar, and filing information. Agents learn through example-based reasoning and are able to improve their performance over time.
2. Agents are computational systems that inhabit some complex, dynamic environment, and sense and act autonomously to realize a set of goals or tasks.
3. Agents are software robots that think and act on behalf of a user to carry out tasks. Agents will help meet the growing need for more functional, flexible, and personal computing and telecommunications systems. Uses for intelligent agents include self-contained tasks, operating semi-autonomously, and communication between the user and systems resources.
4. Agents are software programs that implement user delegation. Agents manage complexity, support user mobility, and lower the entry level for new users. Agents are a design model similar to client-server computing, rather than strictly a technology, program, or product.

Franklin and Graesser [4] have collected and analyzed a more extended list of definitions:

*An agent is anything that can be viewed as perceiving its environment through sensors and acting upon that environment through effectors,* Russel and Norvig, [10].

*Autonomous agents are computational systems that inhabit some complex dynamic environment, sense and act autonomously in this environment. By doing so they realize a set of goals or tasks for which they are designed,* Maes, [8].

*Let us define an agent as a persistent software entity dedicated to a specific purpose. 'Persistent' distinguishes agents from subroutines; agents have their own ideas about how to accomplish tasks, their own agendas. 'Special purpose' distinguishes them from entire multifunction applications; agents are typically much smaller,* Smith, Cypher and Spohrer, [11].

*Intelligent agents continuously perform three functions: perception of dynamic conditions in the environment; action to affect conditions in the environment; and reasoning to interpret perceptions, solve problems, draw inferences, and determine actions,* Hayes-Roth, [5].

*Intelligent agents are software entities that carry out some set of operations on behalf of a user or another program, with some degree of independence or*

*autonomy, and in so doing, employ some knowledge or representation of the user's goals or desires* [6].

*An autonomous agent is a system situated within and a part of an environment that senses that environment and acts on it, over time, in pursuit of its own agenda and so as to effect what it senses in the future,* Franklin and Graesser, [4].

However, as we mentioned in the Introduction, there are also natural and social agents. For instance, the term *"agent"* in the context of business or economic modeling refers to natural real world objects, such as organizations, companies or people. These real world objects are capable of displaying autonomous behavior. They react to external events and are capable of initiating activities and interaction with other objects *(agency)*.

Thus, it is reasonable to assume that an *agent* is anything (or anybody) that can be viewed as perceiving its environment through *sensors* and acting upon this environment through *effectors*. A human agent has eyes, ears, and other organs for sensors, and hands, legs, mouth, and other body parts for effectors. A robotic agent uses cameras, infrared range finders and other sensing devices as sensors and various body parts as effectors. A software agent has communication channels both for sensors and effectors.

This gives us the following informational structure of an agent, reflecting agent's information flows:

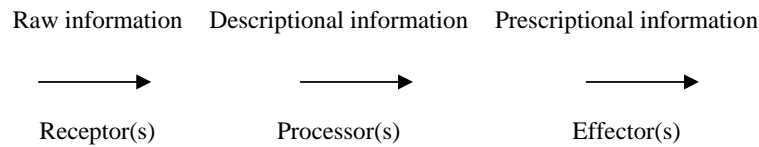

Raw information   Descriptional information   Prescriptional information

Receptor(s)        Processor(s)                Effector(s)

**Fig. 1.** The triadic informational structure of an agent.

## 3 Agent Typology

There are different types of intelligent agents. For instance, Russel and Norvig [10] consider four types:

1. Simple reflex (or tropistic, or behavioristic) agents.
2. Agents that keep track of the world.
3. Goal-based agents.
4. Utility-based agents.

The general structure of the world in the form of the Existential Triad gives us three classes of agents:

- *Physical agents.*
- *Mental agents.*
- *Structural or information agents.*

People, animals, and robots are examples of physical agents. Software agents and *Ego* in the sense of psychoanalysis are examples of mental agents. The head of a Turing machine (cf., for example, Burgin, [3]) is an example of a structural agent.
Physical agents belong to three classes:

- *Biological agents.*
- *Artificial agents.*
- *Hybrid agents.*

People, animals, and microorganisms are examples of biological agents. Robots are examples of artificial agents. Hybrid agents consist of biological and artificial parts (cf., for example, Venda, [14]).

Mizzaro [9] classifies agents according to three parameters. With respect to perception, he identifies *perceiving agents* with various perception levels. *Complete perceiving agents*, who have complete perception of the world, constitute the highest level of perceiving agents. Opposite to perceiving agents, there are *no perception agents*, which are completely isolated from their environment. This does not correlate with the definition of Russell and Norvig [10] but it is consistent with a general definition of an agent.

With respect to reasoning, there are reasoning agents with various reasoning capabilities. Reasoning agents derive new knowledge items from their existing knowledge state. On the highest level of reasoning agents, we have omniscient agents, which are capable of making actual all their potential knowledge by logical reasoning. Opposite to reasoning agents, Mizzaro [9] identifies nonreasoning agents, which are unable to derive new knowledge items from the existing knowledge they possess.

With respect to memory, there are permanent memory agents, which are capable of never losing any portions of their knowledge. No memory agents are opposite to permanent memory agents because they cannot keep their knowledge state. Volatile memory agents are between these two categories. People are volatile memory agents.

Here we suggest seven classifications of agents based on attributive dimensions of agents.

According to the cognitive/intelligence criterion, there are:

1. *Reflex* (or *tropistic*, or *behavioristic*) agents, which realize the simple schema **action → reaction**.
2. *Model based* agents, which have a model of their environment.
3. *Inference based* agents, which use inference in their activity.
4. *Predictive* (*prognostic*, *anticipative*) agents, which use prediction in their activity.

5. *Evaluation based agents*, which use evaluation in their activity.

Some of these classes are also considered in Russell and Norvig [10].

Note that prediction and/or evaluation do not necessarily involve inference.

In addition, according to the dynamic criterion, there are:

1. *static* agents, which do not move (at least, by themselves), e.g., desktop computer;
2. *mobile* agents, which can move to some extent of freedom;
3. *effector mobile* agents, which have effectors that can move;
4. *receptor (sensor) mobile* agents, which have receptors that can move.

Note that mobility can be realized on different levels and in different degrees.

According to the interaction criterion, there are:

1. *deliberative* (*proactive*) agents, which try to anticipate what is going to happen in their environment and to organize their activity taking into account these predictions;
2. *reactive* agents, which react to changes in the environment;
3. *inactive* agents, which do the same thing independently of what happens in the environment.

According to the autonomy criterion, there are:

1. *autonomous* agents;
2. *dependent* agents;
3. *controlled* agents.

There are many kinds and levels of dependence. Control is considered as the highest level of dependence.

According to the learning criterion, there are:

1. *learning* agents do not learn at all;
2. *remembering* agents realize the lowest level of learning – remembering or memorizing
3. *conservative* agents, do not learn at all.

According to the cooperation criterion, there are:

1. *competitive* agents, do not collaborate but only compete;
2. *individualistic* agents, which do not interact with other agents;
3. *collaborative* agents.

There are many kinds and levels of competition. For instance, it can be competition by any cost or competition according to definite (e.g., moral) rules or/and principles. Collaboration also can take different forms.

There are many kinds and levels of dependence. Control is considered as the highest level of dependence.

According to the learning criterion, there are:

1. *learning* agents do not learn at all;
2. *remembering* agents realize the lowest level of learning – remembering or memorizing
3. *conservative* agents, do not learn at all.

There is one more dimension (criterion for classification), which underlies all others. It is the algorithmic dimension. Indeed, an agent can perform operations (e.g., building a model of the environment or making evaluation) and actions (e,g., moving from one place to another) in different modes and using various types of algorithms.

According to the learning criterion, there are:

1. *subrecursive* agents use only subrecursive algorithms, e.g., finite automata;
2. *recursive* agents can use any recursive algorithms, such as Turing machines, random access machines, Kolmogorov algorithms or Minsky machines;
3. *super-recursive* agents can use some super-recursive algorithms, such as inductive Turing machines or trial-and-error machines.

The difference between recursive and super-recursive agents is that at some moment after receiving or formulating a task and starting to fulfill it, a recursive agent will stop and inform that the task is fulfilled. In a similar situation, super-recursive agent can fulfill tasks that do not demand stopping. For instance, a program of a satellite computer which performs observations of changes in the atmosphere for weather prediction has to do observations all time because there is no such a moment when all these observations are completed. As a result, super-recursive agents can perform much more tasks and solve much more problems than recursive agents (cf., for example, [2]).

A *cognitive agent* has a system of knowledge K. Such an agent perceives information from the world and it changes the initial knowledge state, i.e., the state of the system K.

In general, agents may be usefully classified according to the subset of these properties that they enjoy, Franklin and Graesser, [4]. When properties are organized in definite classes, it is possible to use sub-classification schemes via control structures, via environments (database, file system, network, Internet), via language or via applications. For instance, distinction between data-based and knowledge-based agents is based on such part of agent environment as the source of information. Generalizing the approach of Franklin and Graesser [4], we can classify agents by their internal and external components. For instance, control structure is an internal component, while a source of information is an external component. A slightly different approach to the taxonomy of agent properties is based on aspect of an agent, for example, on agent functions. Thus, separation of signal and image analysis agents is related to agent functions. Brustoloni [1] offers another classification by functions: regulation, planning and adaptive agents.

Different types of automata can be associated with the types of agents. Reflex agents may be modeled by automata without memory, such as may be represented by decision tables. All other types of agents demand memory. The third and higher levels, in addition, need a sufficiently powerful processor, varying by the level of the agent. Agents that perform simple tasks may use a finite automaton processor. More sophisticated agents demand processors that perform inference and have the computational power of Turing machines. Processors and program systems for intelligent agents have to utilize super-recursive automata and algorithms, Burgin, [3].

## Conclusions

Agency and agent-based solutions for a wide variety of classes of problems is becoming more and more interesting and important as we face situations of ubiquitous computing, ambient intelligence, autonomous computing, intelligent systems and intelligent robotics – to name but a few issues. In this paper, we presented an extended classification of artificial agents as a contribution to a systematic approach. The aim of this classification is to better understand what kinds of agents can be created and what type of problems demand a specific kind of these agents for their solution. The need for improvement of our basic understanding of what agents are, what they can do and what they can be made to be and to do is very essential for our civilization.